# *Deep Learning Based Natural Language Processing*

Sarvesh Patil

*Dept. Of Electronics and Telecommunication*

*Vivekanand Education Society's Institute of Technology*
Chembur, India
2015sarvesh.patil@ves.ac.in

*Abstract*— Deep Learning methods employ multiple processing layers to learn hierarchial representations of data. They have already been deployed in a humongous number of applications and have produced state-of-the-art results. Recently with the growth in processing power of computers to be able to do high dimensional tensor calculations, Natural Language Processing (NLP) applications have been given a significant boost in terms of efficiency as well as accuracy. In this paper, we will take a look at various signal processing techniques and then application of them to produce a speech-to-text system using Deep Recurrent Neural Networks.

*Keywords—Deep Neural Network, Recurrent Neural Networks, Natural Language Processing, Speech Recognition.*

## I. INTRODUCTION

Natural Language Processing is a method for analysis and representation of human language. NLP methods are heavily theoretical and require a lot of in-depth knowledge of the way signals behave when mathematical modulations are applied to them. NLP in older times used to take up a huge amount of time just to get simple language processing done. Punch Cards and batch processing would take up to 7 minutes to get a simple task done. With the advent of highly efficient algorithms and massive computational power, we can do the same tasks in a matter of seconds. In fact, the Google search engine can scrape up to a million webpages in a matter of milliseconds. NLP enables computers to perform a variety of natural language related tasks at all levels, ranging from parsing and part-of-speech (POS) tagging, to machine translation and dialog systems, to speech recognition.

Deep Learning algorithms are already highly scalable and produce highly accurate results. In fields like Computer Vision or data analytics, they have even surpassed the "Human Level Performance". But we are not quite there yet to make such claims about speech processing and recognition. A very crucial aspect of the research going on in the NLP department is the focus on speech recognition and translation. We can easily make Hidden Markov Models connected in Recurrent fashion work for us in this domain, but they can only achieve so much and can't beat Deep Neural Networks.

The main reason why we choose Deep Networks is that, "end-to-end" tasks can be accomplished by using proper algorithms and training intuitions. End-to-end Neural Networks are the ones which require almost none to very less preprocessing. They learn the entire mapping by themselves provided the network is robust enough. In speech recognition tasks too, such a method can be applied to achieve end-to-end results which are far more accurate than the traditional machine learning approach. Recent work by (Alex Graves et al., 2014) have been the inspiration behind this paper and the Deep Learning techniques behind it.

In section II, we will discuss the mathematical preprocessing required for audio signal processing. In section III, we will take a look at various architectures that RNNs can be composed by and choose the one we will go ahead with. In section IV, we will look at the precautionary steps required before training our network. In section V, we will provide a basic abstract for the training of the network. As this work hasn't been already being done by me, comparative graphs and statistics can't be provided. But in section VI, we will look at the applications this project may help in and finally section VII, will summarize the paper and concluding remarks will be given.

## II. FOURIER ANALYSIS OF SIGNALS

Probably the most important applications of Fourier transforms are pertaining to signal processing. We will cover this topic with an example. Suppose a dilettante plays a piano. When they play a single note on the piano, we hear a certain type of sound. When another note is played, a different type of sound is perceived. Similarly, when a myriad of notes is being played by the pianist, a myriad of sounds is heard by the human ear. Our brains can decipher, with certain training of course, the constituent notes in the symphony being played. But what about computers? A computer, when needs to be shown data of some kind, must be shown digitally. For this purpose, we use a very crucial mathematical tool, invented by Joseph Fourier, the Fourier transform. Fourier Analysis of a signal gives us very important insights about the signal, the constituent frequencies with respect to their amplitudes in the frequency domain.

Let us consider the following example: six different sinusoidal signals with different frequencies are superimposed to form the following signal:

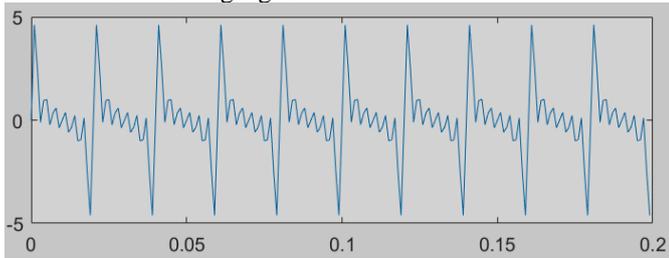
A signal composed of various sinusoidal waves added together

If we take the Fourier transform of this signal, and plot the amplitude of the output variable across various frequencies, we get a graph like this:

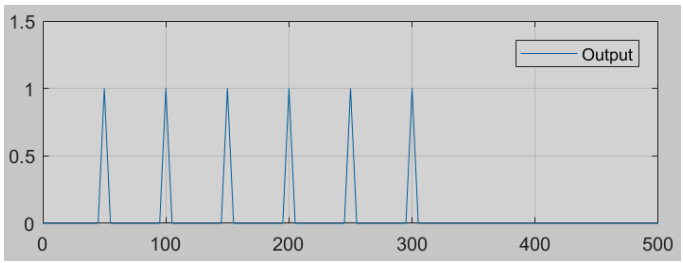

We can now easily conclude that frequencies 50Hz, 100Hz, 150Hz, 200Hz, 250Hz and 300Hz were combined to form the original signal. This process includes 3 steps.
1. Define the constituent signals and add them or use any audio clip and feed it as the input layer.
2. Sample the signal with sampling frequency at least twice that of the maximum frequency in the input signal. In this case, maximum frequency was 300Hz, so a sampling signal with frequency 600Hz and above can be used to convert the continuous signal into discrete form.
3. Apply fast Fourier transform on the discrete signal to generate a function with variable as 'f' which can take values from 0Hz to 500Hz.

Any signal that needs to be pre-processed can be done so, with the aforementioned steps. Once we have a list of all the constituent frequencies in the input signal, for speech recognition, frequencies greater than 4kHz can be neglected as human speech frequencies lie within that range.

### III. RECURRENT NEURAL NETWORKS

RNNs have been in use for quite some time now and in this paper, we are sticking with RNNs because of their dependence on the previous computations and results. This makes sequences in a language intuitive for the Neural Network. There are two most prominent RNN frameworks which provide the best sequence modelling:
1. Long Short-term Memories (LSTMs) (Hochreiter and Schmidhuber, 1997; Gers et al., 1999)
2. Gated Recurrent Units (GRUs) (Cho et al., 2014)

A simple recurrent neural network is shown in Fig. 3.1.

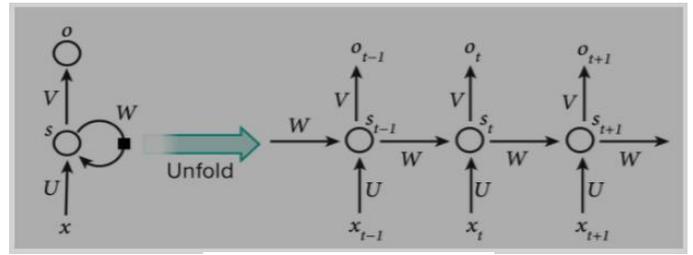
Fig 3.1 (Tom Young et al., 2017)

Each LSTM cell works as shown in Fig 3.2.

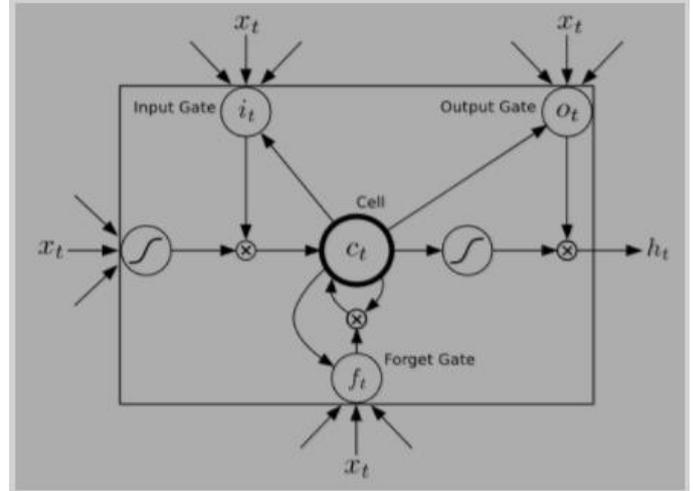
Fig 3.2 LSTM cell (Alex Graves et al., 2014)

In GRUs, each cell 's' looks as shown in Fig 3.3.

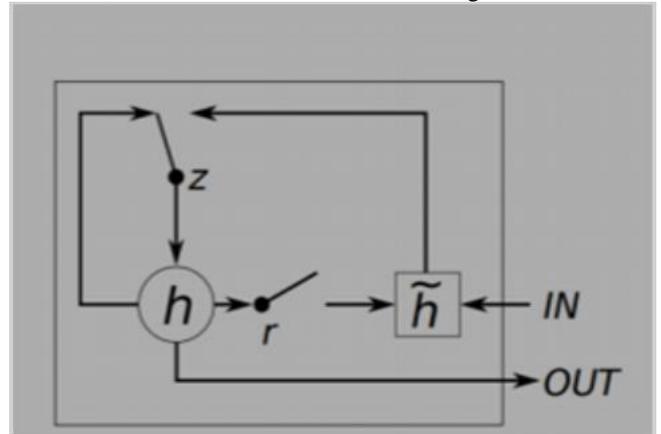
Fig 3.3 GRU gate (Chung et al., 2014)

Output of each LSTM unit can be calculated as:
$i_t = \sigma (W_{xi}x_t + W_{hi}h_{t-1} + W_{ci}c_{t-1} + b_i)$
$f_t = \sigma (W_{xf}x_t + W_{hf}h_{t-1} + W_{cf}c_{t-1} + b_f)$
$c_t = f_t c_{t-1} + i_t * \tanh(W_{xc}x_t + W_{hc}h_{t-1} + b_c)$
$o_t = \sigma (W_{xo}x_t + W_{ho}h_{t-1} + W_{co}c_t + b_o)$
$h_t = o_t * \tanh(c_t)$

where σ is the logistic sigmoid function, and i, f, o and c are respectively the input gate, forget gate, output gate and cell activation vectors, all of which are the same size as the hidden vector h. For our analysis, only LSTMs will be considered because of their high complexity, with sequences with large number of words.

Its unique mechanism enables it to overcome both the vanishing and exploding gradient problem.

For providing input to the first layer for every epoch, we use a Continuous Bag of Words (CBOW). By preprocessing the input, we get sample values at every discrete step in our network. We can use these values by considering 20ms versions of the total information. This will yield a fixed length input of 20ms chunks of data that will be trained upon by the RNN. An example will make the concept clear. Suppose the voice input by a person is "Welcome" and it spans a duration of 700ms. Thus, we will have 30 windows of input data to our RNN. Suppose the network outputs its 30 characters as: <space>__WWWWEEEELLLL_CCCCAAAA__AAAMMMM__<space>. We process the output in the following steps:
1. Separate all the words between the spaces:
   __WWWWEEEELLLL_CCCCAAAA__AAAMMMM__

2. Replace multiple characters by single characters:
   __WEL_CA_AM__
3. Remove the blanks: WELCAAM

Now, as humans we understand 'welcaam' must not be the correct spelling of 'welcome'. To make the kernel understand this, an RNN transducer is used. Generally, a transducer is a device which converts one form of energy into another. Talking digitally, our transducer will convert the acoustically generated output into a linguistically accurate one.

Thus, for detecting entire transcriptions like "She could not be more right", the network may predict "She curd net be more right". Such errors are to be removed only by increasing the training set by feeding the network with more books and news articles. Out of all the predicted transcriptions, we throw out the ones that seem least likely and keep the ones that are more realistic. A possible case of training on a highly morphologically complex language, this character wise embedding becomes exponentially better at prediction than CBOW. Context Specific Vectors from the paper on Character wise embeddings by (Xiaoqing Zheng et al., 2012) have proved to be much more efficient than most other embedding types. CVSs can generate the context representation of a word/character and learn the word/character vector carrying the meaning inferred by that context.

## IV. TRAINING THE NEURAL NETWORK

### A. Connectionist Temporal Classification

LSTMs provide much of the necessary regularization by themselves. They help prevent both, the vanishing as well as the exploding gradient problems. Hence, they're more popular even though they may be computationally greedy than GRUs or CNNs. Connectionist Temporal Classification (CTC) (A. Graves et al., 2006; A. graves et al., 2012) uses a SoftMax layer to define a separate output distribution $Pr(k|t)$ at every step t along the input sequence. The distribution contains K phonemes and two extra symbols, <blank> and <space>. For English, 44 phonemes are defined but can be manipulated to accommodate the phonemes from any other language by changing the value of K.

### B. Bidirectional Recurrent Neural Networks

Bidirectional RNNs are different from conventional RNNs in the aspect that, they calculate the summed output over the previously computed results as well as take into consideration the characters that follow the current node. Fig 4.1 shows the block diagram of a BRNN.

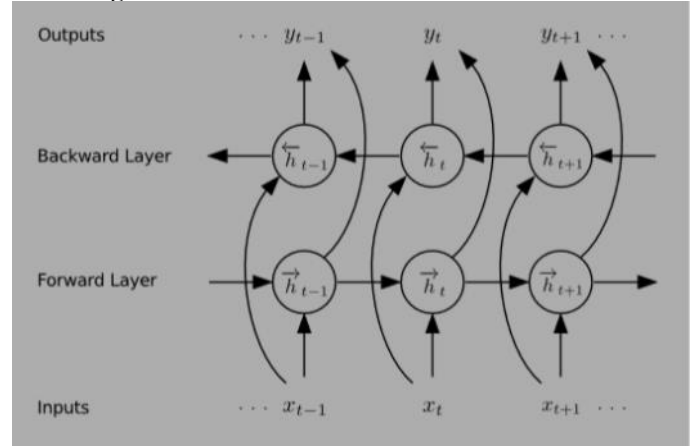

Fig 4.1 BRNN (Alex Graves et al., 2014)

### C. RNN Transducers

CTC defines a distribution over phoneme sequences that depends only on the acoustic input sequence x. It is therefore an acoustic-only model. A recent augmentation, known as an RNN transducer (A. Graves et al., 2012) combines a CTC-like network with a separate RNN that predicts each phoneme given the previous ones, thereby yielding a jointly trained acoustic and language model. The fundamental idea is that, CTCs generate sequences of phonemes
producing an acoustic definition of the input and the RNN transducers will produce a linguistic definition of the word based on the characters. Both the outputs are multiplied to form a common generated text transcription, that will produce a much more accurate result. In particular, Alex Graves et al. Found that the number of deletion errors during decoding is drastically reduced.

### D. Decoding

RNN transducers can be decoded to yield an n-best list of candidate transcriptions. We simply sort the list by their probability values instead of normalized values, as log(Pr)/|Pr| is useful in the case of more deletions than insertions. With deletions reduced to a very small scale, only sorting by Pr will suffice our needs.

### E. Regularization

Regularization techniques help prevent overfitting of the data to the training set. Many methods have been devised to tackle this issue. We find that using dropouts in RNNs is really useful to give a superior performance, provided it is correctly applied. Dropouts in recurrent sections can prove to be fatal to the RNN.

Hence, it's better to be circumspective while applying dropouts to RNNs.

Another very useful regularization technique is addition of weight noise (noise induced in the network weights during training). Weight noise tends to 'simplify' neural networks, in the sense of reducing the amount of information required to send the parameters which improves generalization.

## V. APPLICATIONS

The most promising applications of RNNs remain to be Part of Speech tagging (POS), word level classification, sentence level classification, etc. But what this paper aims at is, speech recognition is very shallow if the computer doesn't understand the context in which the transcription is meant to be. A very critical focus now-a-days is on sentiment analysis, tone analysis and sarcasm detection. Every application requires a new dataset with thousands of examples of real human speech. This is one of the fundamental reasons why the progress has been slow in NLP. Ultimately, end-to-end speech translation is the benchmark that we are yet to conquer, but it won't take us long enough with TPUs coming into play in the near future.

## VI. CONCLUSION

Deep Recurrent Neural Networks coupled with bidirectional Long Short-Term Memory cells have achieved astonishingly good results on various tests around the globe. CTCs are very useful for acoustic modelling of sound and RNN transducers coupled with CTCs can produce semantic relationships with ease while keeping the syntactic relations alive. An obvious step forward would be to actually apply these concepts into such a neural network and then step upon the myriad of pedestals that we are yet to conquer.